# Multi-Object Tracking based on Imaging Radar 3D Object Detection

Patrick Palmer[1], Martin Krüger[1], Dr. Richard Altendorfer[2] and
Univ.-Prof. Dr.-Ing. Prof. h.c. Dr. h.c. Torsten Bertram[1]

[1] TU Dortmund University, Institute of Control Theory and Systems Engineering,
44227 Dortmund, Germany
[2] ZF Group, 56070 Koblenz

**Abstract.**

Effective tracking of surrounding traffic participants allows for an accurate state estimation as a necessary ingredient for prediction of future behavior and therefore adequate planning of the ego vehicle trajectory. One approach for detecting and tracking surrounding traffic participants is the combination of a learning-based object detector with a classical tracking algorithm. Learning based object detectors have been shown to work adequately on lidar and camera data, while learning based object detectors using standard radar data input have proven to be inferior. Recently, with the improvements to radar sensor technology in the form of imaging radars, the object detection performance on radar was greatly improved but is still limited compared to lidar sensors due to the sparsity of the radar point cloud. This presents a unique challenge for the task of multi-object tracking. The tracking algorithm must overcome the limited detection quality while generating consistent tracks. To this end, a comparison between different multi-object tracking methods on imaging radar data is required to investigate its potential for downstream tasks. The work at hand compares multiple approaches and analyzes their limitations when applied to imaging radar data. Furthermore, enhancements to the presented approaches in the form of probabilistic association algorithms are considered for this task.

## 1 Introduction

Building an accurate environment model is of high importance for an automated vehicle. Tracking of traffic participants is a critical component in the automated driving stack since it gives an estimation of the current vehicle state and allows an anticipation of future states. This, in turn, enhances detection performance by maintaining tracks and therefore objects, even when not detected by the sensor. Consequently, it reduces uncertainties in the environment model and serves as the foundation for downstream tasks such as trajectory prediction and motion planning. Current tracking by detection methods consists of two main components: an object detector that estimates object states based on sensor data input and a multi-object tracker.

Various sensor technologies have been utilized for the task of object detection. Lidar sensors are particularly valuable for their high accuracy and resolution, while camera sensors are notable for their cost-effectiveness. For the last 25 years, radar sensors have been employed for object detection task due to their unique benefits. Radar sensors are independent of the lighting conditions, do not suffer significantly from adverse weather conditions, and supply a direct measurement of the relative radial velocity of targets while still being cost-effective. However, until recently, they were limited by low angular resolution, which in turn limited the object detection performance. These issues have been mitigated to a significant extent by the recent development of high-resolution 3+1D imaging radar sensors, which provide a higher point cloud resolution and can additionally measure a target's elevation angle. The work at hand analyzes recently proposed tracking algorithms applied to deep neural network-based (DNN) object detectors and their limitations when applied to imaging radar data. Furthermore, enhancements to the presented approaches, based on parameter optimization and the utilization of probabilistic association metrics, are investigated.

The remainder of the paper is organized as follows. In Section 2 related work is discussed. Section 3 analyzes the utilized tracking methods and describes the probabilistic association metric. Section 4 presents the experiments and discusses the observed outcome. Section 5 concludes the paper and provides an outlook for future work.

## 2    Related Work

Tracking algorithms currently used in neural network-based 3D multi-object tracking can be divided into two main categories. First, joint detection and tracking (JDT) approaches combine object detection and track association tasks into a single end-to-end learnable architecture. This has the advantage of not relying on fixed and pre-defined parameters but parameters that can be adapted through the training process. Fast and Furious [16] and PnPNet [17] describe end-to-end trainable detection, tracking, and motion forecasting approaches that solve the tracking task implicitly. InterTrack [18] utilizes an interaction transformer to aggregate global information from the object detector and performs data association and track management in an end-to-end framework. Tracking-by-detection (TBD) based architectures, as the second main category, employ a two-step process, where an object detector is first utilized to generate an object detection proposal. The multi-object tracker (MOT) subsequently performs tracking and association on the object detections. AB3DMOT [5] proposes a baseline method for MOT utilizing the intersection over union (IoU) for association in conjunction with the Hungarian algorithm (HA) [27] and a Kalman filter for the track prediction. Chiu et al. [4, 31] propose to use the Mahalonobis distance [24] as an association metric instead of the IoU and pairs this with a greedy association instead of HA. SimpleTrack [7] analyzes the limitations of current TBD frameworks and proposes an optimized method that utilizes generalized IoU (GIoU). CenterPoint [8] predicts vehicle motion using a learned motion instead of a motion model and utilizes L2 distance in conjunction with greedy association for association. CasTrack [6] introduces a cost term based on appearance, geometry, and velocity as a similarity metric and matches tracks to detections using a greedy association with dynamic search radius. AB3DMOT [5] and CasTrack [6] show that TBD approaches are sufficiently accurate and can surpass joint detection and tracking frameworks given a sufficiently accurate object detector.

Although object detection and subsequent tracking of traffic participants using lidar point cloud data have been extensively studied, there are few works considering 3+1D imaging radar sensors in this domain. Liu et al. [12] compare three tracking approaches. The first involves a TBD method with point object tracking (POT), which solely considers the center positions of detected objects. In the second approach, a JDT procedure with an extended object tracker (EOT) is utilized, further incorporating the cluster of measurement points surrounding the detected object. Finally, a third approach employs a TBD method with an EOT. All approaches are compared on two datasets, View-of-Delft (VoD) [1] and TJ4DRadset [19]. It is shown that the TBD approach with POT performs best. In contrast, the JDT approach is limited by the conventional clustering method's inability to effectively eliminate dense clutter by the high computational complexity resulting from an excessive number of measurement partition hypotheses. The TBD with EOT is limited by its inadequacy to model the unevenly distributed radar point cloud. Hassan et al. [14] aim to improve the data association problem for a TBD approach under the presence of multiple vehicle classes. It is shown that for 2+1D and 3+1D radar data, the simple consideration of multiple vehicle classes does not improve performance, while the proposed feature extraction network, that extracts additional features from the object's appearance for the tracker, does. Pan et al. [13] describe a JDT framework bespoke for 3+1D radar-based moving object detection and tracking. The framework utilizes a motion estimation and a scene flow estimation module to detect moving objects, which are then associated with tracks using a learnable distance metric.

An accurate object detection is the core dependency for an accurate tracking in the tracking-by-detection approaches. For imaging radar data, this represents a unique challenge, due to the sensors limited resolution in azimuth and elevation. A comparison of state-of-the-art point cloud object detection algorithms shows that voxel and especially pillar-based approaches are well suited for this task [2]. More recently, 3+1D imaging radar-specific models that utilize a pillar representation have been proposed. RPFA-Net [20] extracts global features from the 3+1D radar point cloud using a pillar attention module to create an improved pillar feature representation. Tan et al. [21] combine accumulated temporal and spatial features to perform an interframe association.

## 3   3D Multi-Object Tracking

This section first presents the general tracking pipeline as well as the state-of-the-art TBD approaches that are investigated in this work. Afterwords, the modifications done to optimize for the specific characteristics of radar sensors, as well as the utilized probabilistic association approaches, are described.

## 3.1 Tracking Pipeline

All TBD methods follow a consistent combination of modules, which can be broken down into five main components, given an object detection $D_t$: pre-processing, track prediction, similarity determination, track assignment, and life cycle management. Figure 1 provides a visual representation of the generalized tracking pipeline. The configuration of the components varies, resulting in multiple distinct approaches. Five of these are analyzed and discussed in this work: AB3DMOT [5], AB3DMOT-MH [31], CasTrack [6], SimpleTrack [7], and CenterPoint [8]. Table 1 gives an overview of the differences between the considered approaches.

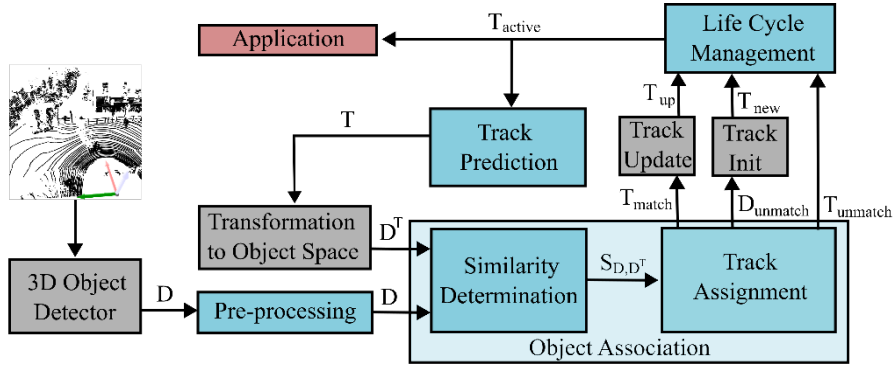

**Figure 1.** Generalized MOT pipeline. An object detector generates predictions D. After a pre-processing step the similarity $S_{D,D^T}$ between predicted track positions $D^T$ and detections D is determined. Detections are either matched to existing tracks $T_{match}$ or remain unmatched $D_{unmatch}$. When no detection fits a track, the track is marked as unmatched $T_{unmatch}$. For unmatched tracks new active trajectories are initiated (Track Init). Matched tracks are updated utilizing a track update with the matched detection state (Track Update). For active tracks the state in the following time step is predicted utilizing the Track Prediction. Active tracks can be utilized in downstream applications, like trajectory prediction.

**Pre-processing**

Objects detected by a 3D object detector are first filtered using a pre-processing step to reduce the number of false positive detections and subsequently tracks. This happens after a non-maximum suppression (NMS) is already utilized in the output stage of most object detectors [9, 22, 23]. AB3DMOT, CASTrack, and CenterPoint utilize score-based thresholding, explicitly utilizing the prediction confidence score (PCS). The disadvantage of PCS-based filtering is that in addition to false detections, true detections with a low PCS will be removed. To mitigate this issue, SimpleTrack introduces the use of NMS filtering at this stage.

**Track Prediction**

A motion forecasting of active tracks from the previous timestep is necessary to connect moving objects between time steps. Most approaches utilize a Kalman filter (KF) [26] to infer vehicle motion. CenterPoint utilizes learned object motion (LoM). The motion is inferred in the CenterPoint object detector's center-head [8]. It, therefore, differs from the other approaches by being the only one that uses some additionally learned parameter in the tracking pipeline.

AB3DMOT, CasTrack, and CenterPoint utilize a constant velocity (CV) in 2D horizontal Cartesian coordinates relative to the ego vehicle, for the state prediction. Although adequate for predicting the current track location when detections are not missed or are only missed for a brief period, this results in a notable position error when an object remains unobserved over an extended duration [6]. CasTrack, therefore, introduces the use of constant acceleration (CA) instead of CV.

**Table 1.** Overview of considered TBD approaches and their attributes: prediction confidence score (PCS), non-maximum suppression (NMS), Kalman filter (KF), learned object motion (LoM), constant velocity (CV), constant acceleration (CA), intersection over union (IoU), generalized IoU (GIoU), Mahalanobis distance (Maha), aggregated pairwise cost (APC), prediction confidence-guided data association (PCGDA), Hungarian algorithm (HA), count-based (CB).

| Approach | Pre-processing | Track Prediction | Similarity | Track Assignment | Life Cycle Management |
|---|---|---|---|---|---|
| AB3DMOT [5] | PCS | KF + CV | IoU | HA | CB |
| AB3DMOT-MH [31] | PCS | KF + CV | Maha | Greedy | CB |
| CasTrack [6] | PCS | KF + CA | APC | PCGDA | CB |
| SimpleTrack [7] | NMS | KF + CV | GIoU | HA | Two Stage |
| CenterPoint [8] | PCS | LoM + CV | L2 | Greedy | CB |

**Similarity Determination**

Next, the active tracks have to be matched to the current detections. This process can be divided into two distinct steps. The first step involves determining the similarity between active tracks and detections, while the second step focuses on associating active tracks to detections. For calculating the similarity, either distance-based metrics, like the L2 distance, utilized by CenterPoint, or overlap base methods, (IoU) and (GIoU) are utilized. Distance-based metrics and the GIoU have the advantage of providing a similarity measurement value greater than zero, even when the track and detection do not overlap [31]. CasTrack introduces an aggregated pairwise cost (APC) term consisting of geometry, appearance, and motion costs. The geometry cost considers the object's position and size, the appearance cost uses features from the deep learning-based detector backbone, and the motion cost associates different motion vectors [6].

**Track Assignment**

Based on the calculated similarity, a matching between tracks and detections is commonly performed, either with the HA or a greedy assignment, which associates the track with the object with the best similarity score. While HA assignment is globally optimal greedy assignment is not, but has been shown to perform equally to HA and is chosen by CenterPoint due to its simplicity [8]. CasTrack varies from this by introducing a prediction confidence-guided data association (PCGDA), which considers the prediction score and APC to flexibly adjust the search range on which a greedy assignment is performed.

**Life Cycle Management**

The task of this module is to manage the birth and death of tracks. A new track is initialized when a detection is not matched to an existing track. Once an active track is no longer matched, the track is terminated. The most prevalent method for deciding if a track should be terminated is a simple count-based (CB) rule utilized by AB3DMOT, CenterPoint, and CasTrack. When a track is not detected for a certain amount of timesteps, it is terminated. SimpleTrack, on the other hand, introduced a two-stage association metric for life cycle management. First, an association is performed utilizing all objects with a high prediction confidence score, identical to the count-based approaches. In the second stage, all the tracks that are not matched in the first stage are kept alive when a prediction with a low confidence score can be matched to the track. This ensures that tracks are kept alive when there is only low confidence in their appearance, which results in fewer object identifier switches [7].

### 3.2 Probabilistic Association

The state prediction of the current track introduces some uncertainty in the track position, attributed to the use of the KF. However, this uncertainty is not considered in the similarity determination process, which utilizes overlap- or distance-based metrics. Consequently, this introduces variability into the matching of detections to tracks. One way of considering the uncertainty is the utilization of Mahalanobis distance [24] as a distance metric in similarity determination. This is employed by Chiu et al. [4, 31] in conjunction with a greedy assignment for track association. The squared Mahalanobis distance between a detection $\mathbf{d}_{t+1}$ and predicted track states $\mathbf{H}\,\hat{\mu}_{t+1}$ is defined as

$$d^2_{Mahalanobis} = (\mathbf{d}_{t+1} - \mathbf{H}\,\hat{\mu}_{t+1})^T \, \mathbf{S}_{t+1}^{-1} \, (\mathbf{d}_{t+1} - \mathbf{H}\,\hat{\mu}_{t+1}). \tag{1}$$

Where the innovation covariance matrix $\mathbf{S}_{t+1}$ represents the uncertainty of a predicted object position

$$\mathbf{S}_{t+1} = \mathbf{H}\,\hat{\Sigma}_{t+1}\,\mathbf{H}^T + \mathbf{R}, \tag{2}$$

$$\hat{\Sigma}_{t+1} = \mathbf{A}\,\Sigma_t\,\mathbf{A}^T + \mathbf{Q}, \tag{3}$$

$$\hat{\mu}_{t+1} = \mathbf{A}\,\mu_t. \tag{4}$$

With $\hat{\mu}_{t+1}$ being the predicted state and $\hat{\Sigma}_{t+1}$ the corresponding predicted state covariance, while $\mu_t$ is the state mean and $\Sigma_t$ the state covariance. $\mathbf{H}$ is the state observation matrix and $\mathbf{A}$ the state transition matrix. $\mathbf{R}$ and $\mathbf{Q}$ are the covariance matrices of the acceleration and observation noise [4, 31].

Altendorfer et al. [3] analyzed and described the limitations of the Mahalanobis distance when utilized in multi-object tracking. One of the significant limitations being, that detections, which have a sizeable mean offset to the actual object and a significant covariance, have a small Mahalobis distance and might therefore be matched to a track, even when a measurement with smaller mean offset and variance is present. The association log-likelihood (A-LL) distance is derived based on the joint association probability for multi object tracking [29], to mitigate the limitations of the Mahalanobis distance

$$d_{A-LL}^2 = d_{Mahalanobis}^2 + \ln|\mathbf{S}_{t+1}^{-1}| + n\ln(2\pi) - 2\ln(P_D). \tag{5}$$

The first term is the Mahalanobis distance. $n$ is the dimensionality of the measurement state $n = \dim(\mathbf{o}_{t+1} - \mathbf{H}\hat{\mu}_{t+1})$. $P_D$ is the probability of getting a true positive detection for a track given the predicted object $\mathbf{d}_{t+1}$. For a radar point the detection probability could be a function of estimated amplitude or signal-to-noise ratio [3]. Due to $P_D$ and $\mathbf{R}$ being unknown for the utilized dataset and object detection method in this work, it is assumed to be constant.

## 4 Experiments

This section evaluates the described approaches and modifications that were made in regard to probabilistic matching. All experiments are conducted using annotated object positions, lidar detections, and radar detections. Annotated object positions enable the assessment of maximum performance with optimal object detection. Lidar detections enable evaluation using a sensor with lower error rate and high-quality detections.

### 4.1 Dataset

For the experiments, we utilize the View-of-Delft (VoD) dataset [1]. It includes lidar, camera, 3+1D imaging radar sensor data, and 3D bounding box annotations for various traffic participants. Only cars, pedestrians, and cyclists are considered in this work. All scenarios are recorded during midday under clear weather conditions. Due to the withholding of the labels for the original test set, the original validation dataset is utilized as a test dataset in this work. The tracking algorithms are evaluated only on this set. For training of the object detectors, the original training datasets is split into a training and a validation dataset. We utilize an accumulated point cloud spanning over 5 frames for the radar detection, as it has been demonstrated to improve detection performance [1, 25]. The recorded traffic scenario focuses on inner city traffic at low velocities of up to 30 km/h with vulnerable road users like pedestrians and cyclists. Therefore, the provided sensor data and labeled objects are limited to a range of 52.6m.

### 4.2 Evaluation Metrics

Several evaluation metrics for 3D MOT have been proposed in the literature. The most widely used metrics are multi-object tracking accuracy (MOTA) and multi-object tracking precision (MOTP) [11]

$$\text{MOTA} = 1 - \frac{\text{FP}_T + \text{FN}_T + \text{IDS}}{\text{num}_{gt}}, \tag{6}$$

$$\text{MOTP} = \frac{\sum_i d_i}{c}. \tag{7}$$

which depends on the number of the false positive tracks ($FP_T$), false negative tracks ($FN_T$), object identifier switches (IDS), and number of ground truth objects ($\text{num}_{gt}$). The MOTP depends on the number of matches $c$ and the overlap $d_i$ of each target $i$ with its corresponding ground truth object. The interpretation of the MOTA metric is limited by its strict upper bound of less than 1 and no lower bound [5]. For evaluation of the quality of object detections, specifically how often an object is detected correctly, the detection accuracy (DetA) [10] is utilized

$$\text{DetA} = \frac{|TP_D|}{|TP_D| + |FN_D| + |FP_D|}, \tag{8}$$

which depends on the number of true positive ($TP_D$), false positive ($FP_D$) and false negative ($FN_D$) object detections. To assess the quality of track consistency, specifically how many different tracks are predicted for a singular object, the association accuracy (AssA) [10] can be used

$$\text{AssA} = \sum_{b \in \{TP_D\}} \frac{|TPA(b)|}{|TPA(b)| + |FNA(b)| + |FPA(b)|}. \tag{9}$$

It depends on the true positive associations (TPA) false positive associations (FPA) and false negative associations (FNA) for all $TP_D$. These two metrics are combined by the higher order tracking accuracy (HOTA) metric [10], which allows a joint evaluation of detection accuracy and association accuracy. The HOTA is the main metric considered in this work for comparing the model performance

$$\text{HOTA} = \sqrt{\text{DetA} \cdot \text{AssA}}. \tag{10}$$

### 4.3 Object Detection and Ego Motion Consideration

PointPillars [9] is utilized as an object detector, as it produces good object detection performance on 3+1D imaging radar data, compared to other state-of-the-art object detection approaches [1, 2] while still being sufficient on lidar data [2]. For all experiments, the implementation of PointPillars in the OpenPCDet [15] framework is utilized. As shown in Table 2, this approach can detect roughly half of all cars and pedestrians, when used with the lidar sensor data, while more than 65% of bikes are detected correctly. For detections from the radar sensor, it can be observed that while roughly a third of cars and cyclists are detected correctly, only 12.51% of pedestrians are detected correctly. This is due to the high noise of the radar targets and a relatively low number of targets per pedestrian, making them hard to detect. This directly influences the tracking performance. One major limitation of the VoD dataset is the accuracy of the provided ego motion [25]. This can influence the tracking performance due to being utilized in the state prediction. To analyze the effect that the ego motion estimation has on the tracking performance the labeled ego motion is compared to two other derivations of the ego motion.

**Table 2.** Object detection results on the VoD lidar and radar point cloud utilizing the PointPillars architecture, quantified as mean average precision (mAP) in the 3D and bird eye view (BEV) space. Evaluated at an IoU of [0.7, 0.5, 0.5] in [x, y, z] direction for the car class and [0.5, 0.5, 0.5] for the pedestrian and cyclist class. The best results are marked in **bold**.

| Sensor | mAP-3D [%] | | | mAP-BEV [%] | | |
|---|---|---|---|---|---|---|
| | Car | Pedestrian | Cyclist | Car | Pedestrian | Cyclist |
| Lidar | **49.95** | **50.81** | **65.86** | **53.09** | **56.89** | **68.32** |
| Radar | 36.28 | 12.51 | 30.76 | 44.54 | 22.03 | 53.59 |

First a static ego vehicle assumption and second an estimated ego motion which is derived by applying a simultaneous localization and mapping (SLAM) motion estimation to the lidar point cloud sequence [32]. Table 3 lists the results. It can be observed that the difference between tracking utilizing the static ego motion assumption and the labeled motion is only 0.69 percentage points for the HOTA metric. This shows that the tracking algorithm is relatively independent of the ego motion as long as the variance of the motion is low. A contributing factor lies in the capability of the KF to estimate the relative motion between objects and ego vehicle for each track. However, this estimation is only accurate when the variance is low. The SLAM estimated motion has a higher variance between frames than the labeled motion; here, the KF is not able to estimate the speed of an object with sufficient accuracy. This results in a high number of IDSs and false assignments of objects to tracks. Another contributing factor is the dataset's low maximum ego velocity of approximately 30km/h, keeping the influence of ego motion on the objects low. These findings remain consistent when evaluated on the lidar and radar detected objects. Therefore, all further experiments utilize the labeled ego motion.

**Table 3.** Tracking results utilizing different ego motion estimations on the AB3DMOT tracking approach utilizing the labeled object positions. The best results are marked in **bold**.

| Ego Motion Estimation | HOTA [%] | DetA [%] | AssA [%] | MOTA [%] | MOTP [%] |
|---|---|---|---|---|---|
| Labeled Ego Motion | **89.8** | **90.30** | **89.31** | **90.02** | **100** |
| Stationary Ego Vehicle | 89.11 | 90.12 | 88.12 | 89.80 | 100 |
| SLAM Estimation | 61.25 | 59.28 | 63.36 | 58.08 | 100 |

### 4.4 Comparison of State-of-the-Art Architectures

Table 4 shows the quantitative results of tracking objects on labeled object positions, lidar detections, and radar detections with the four considered approaches. First, evaluating the tracking applied to the labeled object position, it can be observed that tracks get missed even for a perfect detection of objects. This can be attributed to the uncertainty of the ego motion, as shown in Section 4.3, and the uncertainty in the labeled object position [25]. Additionally, some objects move out of the visible frame and get back into it again at a later stage. The object label is not present due to being outside of the observable space. In this case, tracking relies on keeping the tracks alive through life cycle management and the track prediction. The matching between a lost track and detection is especially challenging for the AB3DMOT approach whose tracking performance on the labeled object position is 7.45, respectively 8.98 percentage points worse than CasTrack and SimpleTrack. This behavior is also observed on the lidar and radar detections. AB3DMOT fails to assign objects to tracks when they are not detected for multiple frames. It is best demonstrated in a qualitative analysis. Figure 2 to Figure 5 show two frames from the VoD dataset, with detected bounding boxes and track IDs for each of the four approaches overlayed. The coloring of the bounding boxes is random and chosen so that it is easy to distinguish close objects from one another The numbers at each bounding box are the tracking IDs assigned by each tracking approach. Additionally, the radar point cloud is overlaid as blue dots. The pedestrian with the stroller on the far right of the image is observed in frame 236 and frame 242 but not in the frames in between. Despite not detecting this pedestrian for four frames, it is still matched to the same track by CasTrack (Figure 3a and Figure 3b) and SimpleTrack (Figure 4a and Figure 4b), but AB3DMOT initialized a new track, as observed comparing Figure 2a and Figure 2b. The main difference for this case between SimpleTrack and AB3DMOT is the utilization of GIoU instead of IoU as a matching metric, which improves upon the IoU by still providing a meaningful distance metric even if there is no overlap between detection and track. The NMS-based pre-processing in SimpleTrack results in a low detection accuracy score for lidar and radar data. Due to NMS being already used in the output stage of the utilized object detector, no further objects are filtered in this step. This results in many false positive detections, which lead to a low DetA score. Qualitatively, this is observable when comparing Figure 2a. to Figure 4a. The objects with the IDs 654 and 656 are false positives. Furthermore, it results in more correctly tracked objects due to true positive detections, with low con

**Table 4.** Results of the 4 utilized tracking approaches, on the three different detection types. The best results for each detection type are marked in **bold**.

| Approach | Detections | HOTA [%] | DetA [%] | AssA [%] | MOTA [%] | MOTP [%] |
|---|---|---|---|---|---|---|
| AB3DMOT | Label | 89.80 | 90.31 | 89.31 | 90.02 | **100.00** |
| | Lidar | 40.28 | 35.52 | 46.95 | **31.49** | **79.19** |
| | Radar | 34.32 | 27.66 | 43.15 | **18.89** | 72.74 |
| CasTrack | Label | 97.25 | 98.86 | 95.69 | 98.39 | **100.00** |
| | Lidar | **44.80** | **36.01** | 56.03 | 26.94 | 78.64 |
| | Radar | **40.10** | **33.04** | 48.66 | 9.55 | 72.47 |
| SimpleTrack | Label | **98.78** | 99.10 | 98.44 | 99.03 | 100.00 |
| | Lidar | 42.60 | 30.31 | **60.64** | -13.78 | 78.06 |
| | Radar | 33.74 | 23.53 | **48.80** | -32.06 | 72.00 |
| CenterPoint | Label | 96.76 | **99.10** | 94.58 | 97.26 | **100.00** |
| | Lidar | 42.62 | 32.31 | 60.29 | -6.51 | 78.25 |
| | Radar | 37.32 | 29.16 | 48.23 | 15.37 | **72.84** |

Three main factors differentiate radar detections from lidar detections. One is the larger number of false positives that only appear in single frames. In contrast, false positive detections from the lidar sensor remain more consistent over multiple frames. This leads to early termination of tracks when a false positive detection is incorrectly assigned to a track. Another factor is the larger variance in position error of correctly detected objects, which results in an ID switch when multiple objects are close to each other. SimpleTrack suffers the most from this effect, demonstrated by the low HOTA score of SimpleTrack on Radar data. Finally, temporal gaps in radar detections are larger than in lidar detections, which makes the association between tracks and objects harder. This is especially affecting AB3DMOT and CenterPoint. CasTrack is the most robust against these effects, represented by the best performance on the radar HOTA score, compared to the other considered approaches. CasTrack also keeps the object tracks more consistent when high detection uncertainty exists. Overall, considering the tracking on radar and lidar data, CasTrack is the best of the four compared architectures.

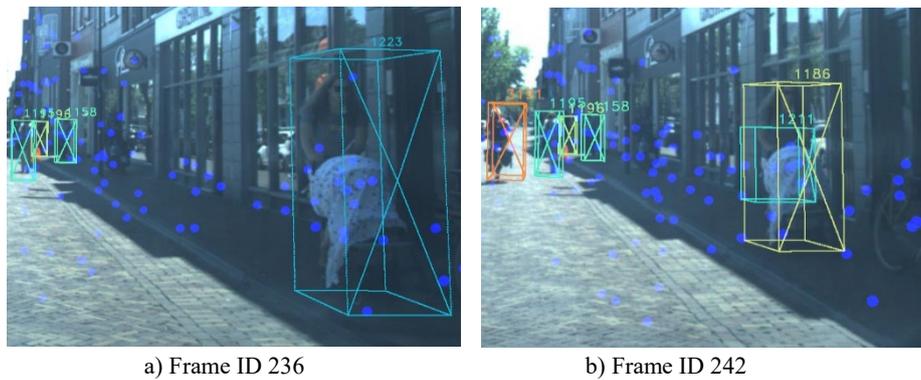

a) Frame ID 236    b) Frame ID 242

**Figure 2.** Qualitative results of AB3DMOT on radar detections. Blue points represent radar reflections.

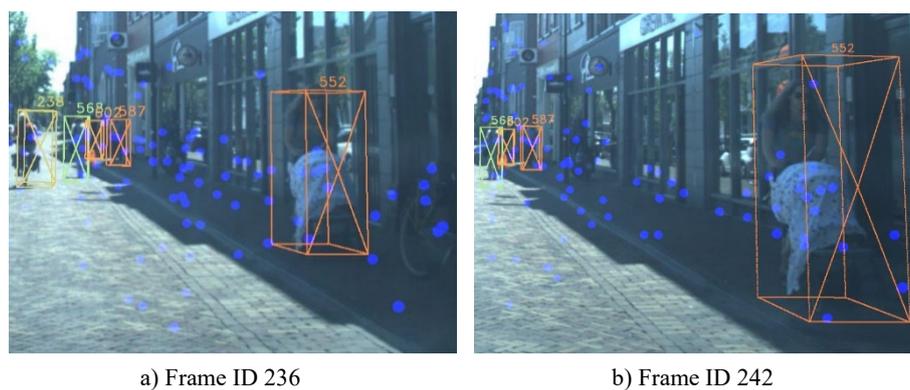

a) Frame ID 236    b) Frame ID 242

**Figure 3.** Qualitative results of CasTrack on radar detections. Blue points represent radar reflections.

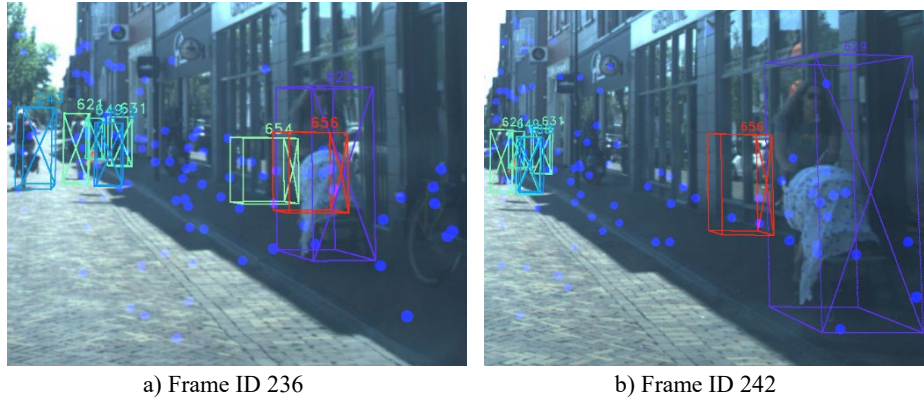

a) Frame ID 236             b) Frame ID 242

**Figure 4.** Qualitative results of SimpleTrack on radar detections. Blue points represent radar reflections.

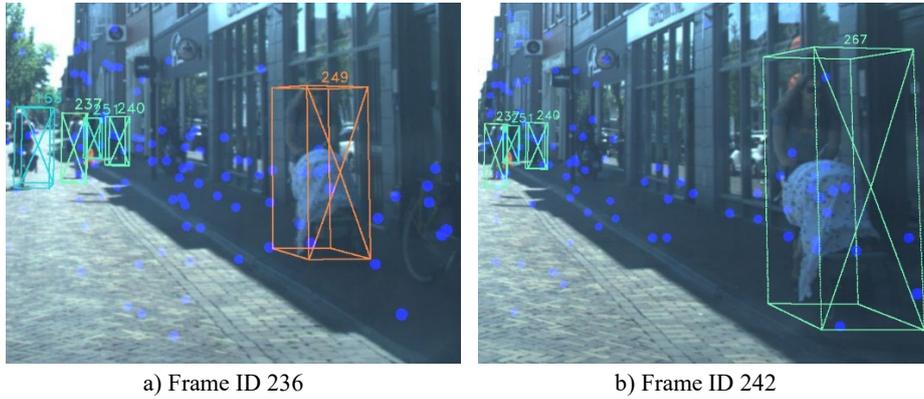

a) Frame ID 236             b) Frame ID 242

**Figure 5.** Qualitative results of CenterPoint on radar detections. Blue points represent radar reflections.

### 4.5 Influence of Probabilistic Association

In this section we explore the effect of probabilistic association distances on the overall tracking performance characterized by tracking metrics. We compare IoU as often used in DNNs as a base distance for association to the Mahalanobis distance, a commonly used heuristic in model based probabilistic tracking approaches, and to the association log-likelihood distance. The other modules are configured as described by AB3DMOT-MH. Experiments are only executed on the radar dataset due to the higher measurement noise compared to lidar and, therefore, the greatest possible benefit when utilizing a probabilistic association metric. Table 5 shows that tracking utilizing the Mahalanobis distance results in a worse outcome for all considered metrics, excluding the MOTP on the car class. By employing the Mahalanobis distance, active tracks frequently assign false positive detections to their trajectory when no true positive object detection is present. An example of this effect is demonstrated in Figure 6. The pedestrian on the far-right side (tracking ID 562) is correctly detected in Figure 6a. In Figure 6b, it is not detected, but a false positive detection is assigned to the track. These false assigned detections are continued over multiple frames. Which results in an IDS for this object once it is again correctly detected in a later frame. This worse assignment contributes to a worse DetA due to false positive objects being kept alive falsely. The higher MOTP might be due to tracks being kept alive when no true positive detection is present. The same effect is observable but less pronounced when utilizing the A-LL distance.

However, this evaluation is limited since the full power of the Mahalanobis and A-LL distance could not be exploited due to the non-availability of the measurement uncertainty and probability of detection from the utilized object detector. As expected, the A-LL performs in many cases better than the Mahalanobis distance, even though the measurement uncertainty and $P_D$ are treated as fixed parameters.

**Table 5.** Tracking results on the objects detected upon radar data utilizing different probabilistic and non-probabilistic association metrics. Results are separated into the 3 considered vehicle classes. The class wise best result is highlighted in **bold**.

| Association Metric | Vehicle Class | HOTA [%] | DetA [%] | AssA [%] | MOTA [%] | MOTP [%] |
|---|---|---|---|---|---|---|
| IoU | Car | **39.92** | **30.81** | **51.87** | **29.95** | 78.33 |
| | Pedestrian | **21.67** | 14.92 | **31.79** | **2.72** | **69.51** |
| | Cyclist | **42.14** | **36.25** | **49.62** | **28.52** | **70.92** |
| Mahalanobis distance | Car | 30.84 | 28.81 | 33.26 | 27.69 | **78.69** |
| | Pedestrian | 18.73 | 14.44 | 24.64 | -2.40 | 69.08 |
| | Cyclist | 34.75 | 34.98 | 35.56 | 24.40 | 70.76 |
| A-LL distance | Car | 38.25 | 30.46 | 48.23 | 28.39 | 78.54 |
| | Pedestrian | 18.72 | **15.85** | 22.61 | -22.94 | 68.53 |
| | Cyclist | 37.81 | 31.91 | 45.55 | 3.07 | 70.62 |

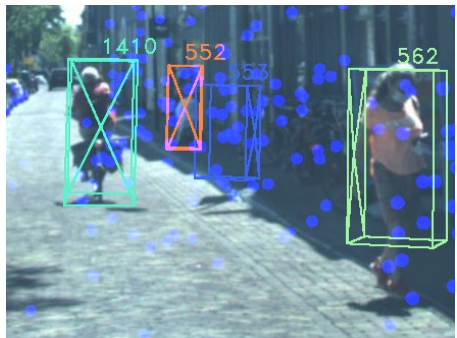 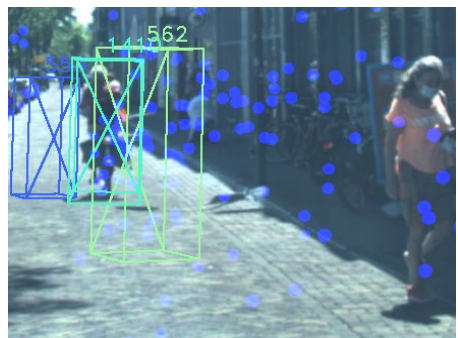

a) Frame ID 190          b) Frame ID 191

**Figure 6.** Qualitative results of probabilistic association utilizing the Mahalanobis distance. Blue points represent radar reflections.

## 5  Conclusion and Outlook

Accurate tracking of traffic participants is a crucial component in the automated driving stack. In the contribution at hand, five recently proposed TBD approaches are compared regarding the tracking performance of objects detected using radar or lidar sensor data from the VoD dataset. Furthermore, alongside the state-of-the-art methods, extensions incorporating the Mahalanobis distance and the A-LL distance as probabilistic association metrics are examined and evaluated on the radar detections.

It is shown that especially AB3DMOT is limited by objects that are not visible for multiple frames; this is already observable when the labeled object positions are considered for tracking where the HOTA score is only at 89.80%, compared to the other considered approaches, which have a score of at least 96.76%. SimpleTrack, on the other hand, suffers from many false positive detections and tracks caused by fewer false positive detections being filtered out in the pre-processing due to the non-utilization of PCS. Generally, CasTrack performs the best out of all the considered approaches with respect to the HOTA and DetA metric due to being the most robust against false positive detections. This is especially noticeable on the radar detections.

The use of probabilistic association metrics for the similarity determination does not seem to improve the tracking performance, for example, due to active tracks being matched to false positive detections, even when there is a high offset to the true object position. However, their influence is limited by the used non-probabilistic object detectors which do not provide measurement uncertainties and probability of detection.

Future work can explore two possible directions: improving the object detection performance or improving the tracking approach. The object detector can be improved either by considering additional sensor types like cameras in conjunction with the radar sensor or by tackling the problem of low radar point cloud resolution by investigating densification methods, which are already applied to sparse lidar data [28]. The tracking algorithms can be improved, by explicitly considering the limitations observed on lidar and radar data. For tracking on radar data, one of the main limitations is the high number of false positive detections, which mainly result in false positive tracks or IDSs of true tracks. This could be tackled by employing a by a better filtering of the radar detections. An advantage of the radar sensor, which should be considered in the track prediction is the direct measurement of the radial velocity. This can either be directly imported into the Kalman filter as an additional input or can be used to cross check the estimated motion against the measurement. An improvement that would have benefits both on detection and tracking level would be the extension of the object detector to a Bayesian detector, in the simplest case by adding uncertainty heads [30]. This also would allow a proper use of the probabilistic association distance.